\documentclass[sigconf,nonacm]{acmart}
\AtBeginDocument{%
  }

\copyrightyear{2025}
\acmYear{2025}
\setcopyright{acmlicensed}\acmConference[WWW Companion '25]{Companion
Proceedings of the ACM Web Conference 2025}{April 28-May 2, 2025}{Sydney,
NSW, Australia}
\acmBooktitle{Companion Proceedings of the ACM Web Conference 2025 (WWW
Companion '25), April 28-May 2, 2025, Sydney, NSW, Australia}
\acmDOI{10.1145/3701716.3717649}
\acmISBN{979-8-4007-1331-6/2025/04}

\usepackage{amsmath}
\usepackage{enumitem}
\usepackage{hyperref}
\usepackage{xcolor}
\usepackage{cleveref}
\usepackage{multirow}
\usepackage{array}
\usepackage[linesnumbered,ruled,vlined]{algorithm2e}

\usepackage{flushend}
\usepackage{balance}

\acmYear{2025}
\acmISBN{}
\acmDOI{}

\title{Blockchain-based Framework for Scalable and Incentivized Federated Learning}

\author{Bijun Wu}
\affiliation{%
  \institution{Rensselaer Polytechnic Institute}
  \city{Troy}
  \country{USA}
}
\email{wub8@rpi.edu}

\author{Oshani Seneviratne}
\orcid{0000-0001-8518-917X}
\affiliation{%
  \institution{Rensselaer Polytechnic Institute}
  \city{Troy}
  \state{NY}
  \country{USA}
}
\email{senevo@rpi.edu}

\renewcommand{\shortauthors}{Wu and Seneviratne}
\renewcommand{\shorttitle}{Scalable and Incentivized Federated Learning}

\begin{document}

\begin{abstract}

Federated Learning (FL) enables collaborative model training without sharing raw data, preserving privacy while harnessing distributed datasets. However, traditional FL systems often rely on centralized aggregating mechanisms, introducing trust issues, single points of failure, and limited mechanisms for incentivizing meaningful client contributions. These challenges are exacerbated as FL scales to train resource-intensive models, such as large language models (LLMs), requiring scalable, decentralized solutions.
This paper presents a blockchain-based FL framework that addresses these limitations by integrating smart contracts and a novel hybrid incentive mechanism. The framework automates critical FL tasks, including client registration, update validation, reward distribution, and maintaining a transparent global state. The hybrid incentive mechanism combines on-chain alignment-based rewards, off-chain fairness checks, and consistency multipliers to ensure fairness, transparency, and sustained engagement. We evaluate the framework through gas cost analysis, demonstrating its feasibility for different scales of federated learning scenarios.

\end{abstract}



\maketitle

\section{Introduction}
\label{sec:introduction}

Federated Learning (FL)~\cite{mcmahan2017communication} has emerged as a transformative paradigm for distributed machine learning, enabling clients to collaboratively train models without sharing raw data. By preserving data privacy and supporting collaborative training, FL has shown promise for applications across diverse domains~\cite{yang2019federated}. However, traditional FL systems face critical limitations that hinder their scalability, reliability, and fairness~\cite{kairouz2021advances,li2020federated}.

First, most FL systems rely on centralized mechanisms to aggregate and validate contributions. This centralization introduces trust issues, as clients must rely on the centralized aggregator to act without bias, manipulation, or misuse of data~\cite{zhang2024}. Second, centralized architectures create single points of failure, making the system vulnerable to outages, attacks, and bottlenecks that compromise reliability~\cite{sani2024}. Finally, as FL systems scale, particularly with the growing adoption of foundation models, existing approaches lack robust incentive mechanisms to encourage meaningful contributions or penalize malicious or low-quality updates that could lead to fairness and performance degradation over time.

These challenges highlight the need for decentralized and transparent FL systems that enhance trust, resilience, and incentivization. 

Inspired by Swarm Learning~\cite{hp_swarm_learning}, a decentralized machine learning framework, our work draws on blockchain-based coordination and tamper-proof state synchronization. Swarm Learning replaces the centralized aggregation mechanism with decentralized nodes and integrates blockchain technology to ensure transparent and immutable synchronization. Trusted execution environments (TEEs) are used to secure data and model parameters. However, Swarm Learning lacks explicit incentive mechanisms to ensure meaningful contributions, limiting its applicability in competitive or heterogeneous environments.

\subsection*{Contributions}
We propose a blockchain-based FL framework that integrates a novel hybrid incentive mechanism. Our framework utilizes smart contracts~\cite{buterin2014next} to automate critical FL tasks, including client registration, update validation, reward distribution, and maintenance of a transparent global state. A hybrid incentive mechanism ensures fairness and scalability by combining:
\begin{enumerate}
    \item \textbf{On-Chain Alignment-Based Rewards:} Evaluate client contributions in real-time to promote high-quality updates.
    \item \textbf{Off-Chain Fairness Checks:} Leverage decentralized storage to ensure equitable reward distribution over time while minimizing blockchain costs.
    \item \textbf{Consistency Multipliers:} Reward sustained, high-quality participation across multiple rounds for long-term engagement.
\end{enumerate}

Our contributions address key challenges in trust, scalability, and incentivization, paving the way for scalable FL systems capable of handling resource-intensive applications. Notably, training Large Language Models (LLMs) demands substantial computational resources—a challenge that is further amplified in decentralized settings, where blockchain constraints such as limited on-chain capacity and high gas costs complicate resource management. Our framework is designed to mitigate these issues by efficiently balancing off-chain and on-chain operations.

\medskip

The rest of the paper is organized as follows. In \Cref{sec:related_work}, we review related work in blockchain-based FL, highlighting existing gaps and complementary approaches. \Cref{sec:architecture} describes the proposed system architecture, focusing on the integration of blockchain and FL. \Cref{sec:incentive_mechanism} details the hybrid incentive mechanism design, including its components and operational flow. In \Cref{sec:evaluation}, we present empirical evaluations of gas consumption under varying model parameter sizes, demonstrating the scalability and efficiency of the framework. \Cref{sec:discussion} outlines limitations of the current work and several future directions. Finally, \Cref{sec:conclusion} concludes the paper with a short summary.

\section{Related Work}
\label{sec:related_work}

Blockchain-based approaches to FL have garnered significant attention for their potential to enhance decentralization, transparency, and trust in decentralized learning systems. Our work aligns with the growing research at the intersection of blockchain and FL~\cite{issa2023blockchain}, focusing on leveraging blockchain to enhance scalability and fairness, particularly for resource-intensive applications like training LLMs. Below, we discuss some of the closest related works to the work presented in this paper.


The Blockchain-Based Decentralized Federated Learning Framework with Committee Consensus (BFLC)~\cite{li2020blockchain} eliminates reliance on a central server by using blockchain for global model storage and local model update exchange. The committee consensus mechanism reduces computational overhead and mitigate malicious attacks. In contrast, our framework employs a hybrid incentive mechanism to ensure fairness and scalability. It addresses challenges such as the high computational demands of large-scale models in decentralized settings by leveraging off-chain aggregation alongside efficient on-chain operations.


The Differentially Private Blockchain-Based Vertical Federated Learning (DP-BBVFL) framework~\cite{tran2024adifferentially} introduces differential privacy to protect embeddings stored on the blockchain, ensuring privacy in vertical FL scenarios. Unlike DP-BBVFL, which focuses on embedding aggregation in settings with disjoint feature spaces, our work targets horizontal FL, where clients share data with the same feature space but different sample distributions. By aggregating model updates with a hybrid incentive mechanism, our approach ensures fairness and promotes meaningful contributions across participants.


Kang et al.~\cite{kang2019incentive} introduce a reputation-based worker selection scheme using a multiweight subjective logic model to evaluate reliability and trustworthiness. Their blockchain-based incentive mechanism integrates reputation with contract theory to motivate high-quality participation. While this approach focuses on trust and reputation, our framework emphasizes the scalability of FL, balancing alignment-based rewards and fairness checks, and enabling efficient training for large-scale models.


The Blockchain and Federated Learning for Privacy-Preserved Data Sharing in Industrial IoT framework~\cite{lu2019blockchain} integrates FL into the consensus process of a permissioned blockchain, allowing computational resources used for consensus to contribute to model training. While this industrial IoT-focused approach addresses privacy and resource utilization, our framework is designed for FL, emphasizing scalability and fairness through hybrid incentives and off-chain processing, making it suitable for diverse applications beyond industrial settings.


DeepChain~\cite{weng2019deepchain} introduces a blockchain-integrated framework for distributed deep learning, employing a protocol-level incentive mechanism to enforce correct participant behavior and mitigate malicious attacks. Unlike DeepChain’s protocol-level integration, our work focuses on application-level hybrid incentives, addressing fairness and scalability in federated training for resource-intensive models.

The Shareable Updatable Model (SUM) framework~\cite{harris2019decentralized,harris2020analysis} proposes a decentralized methodology for collaboratively building datasets and hosting models on public blockchains, employing financial and gamified incentives. While SUM focuses on public blockchains and gamified incentives for collaborative construction of models, our framework emphasizes federated techniques for training models, offering scalability and fairness in heterogeneous environments.

Finally, the Swarm Learning  framework~\cite{hp_swarm_learning} utilizes blockchain-based peer-to-peer networking for decentralized machine learning, ensuring privacy by keeping raw data localized and complying with regulations. While swarm learning achieves transparency and equitable collaboration, it lacks explicit incentive mechanisms to ensure meaningful contributions or penalize malicious behavior, a gap that our hybrid incentive mechanism addresses by combining alignment-based rewards and fairness checks.

By addressing the scalability challenges of FL and integrating a hybrid incentive mechanism, our framework advances the capabilities of blockchain-based FL for large-scale, resource-intensive applications, distinguishing itself from prior works focused on specific FL paradigms or application domains.

\section{System Architecture}
\label{sec:architecture}

\begin{figure}[t]
    \centering
    \includegraphics[width=\linewidth]{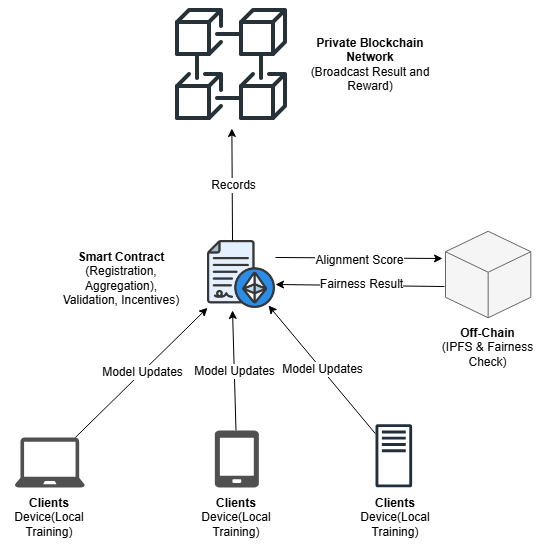}
    \Description{The figure illustrates the interaction between clients performing local training, the smart contract for registration, aggregation, validation, and incentives, the private Ethereum-based blockchain network for disseminating model update results and eventually the rewards to the clients participating in the training process, and the off-chain storage for fairness checks and alignment score computations.}
    \caption{System Architecture.}
    \label{fig:system_architecture}
\end{figure}

The proposed system architecture integrates blockchain technology with FL to create a decentralized and transparent model update framework. 
\Cref{fig:system_architecture} illustrates the interactions within the system, including clients performing local training, the smart contract managing registration, aggregation, validation, and incentives, and the private Ethereum-based blockchain network disseminating model updates and distributing rewards. Additionally, the framework leverages the Interplanetary File System (IPFS) for storing the alignment score computations for efficient fairness checks to ensure transparency and equitable participation in the training process.

The architecture comprises three interdependent components that together facilitate decentralized training, validation, and incentive distribution:

\begin{itemize}
\item \textbf{Private Blockchain Network:}
A private Ethereum-based blockchain forms the backbone of the system, ensuring immutable and transparent recording of critical data such as client contributions, rewards, and fairness evaluations. We employ a private blockchain as we envision such collaborative training taking place in a consortium-based environment, where participants belong to a trusted group. 
Deploying the system on a private blockchain ensures controlled access, reduces transaction fees, and minimizes latency compared to public blockchain networks.
However, the proposed methodology is flexible and can also be adapted for public, permissionless blockchain settings if required. 

\item \textbf{Smart Contracts:}  
Smart contracts automate key processes, ensuring trustless interactions between participants. They handle the following tasks:
\begin{itemize}
    \item \textbf{Registration and Staking:} Clients register on the blockchain and stake tokens as a commitment to meaningful participation.
    \item \textbf{Update Validation:} Submitted updates are validated for alignment with the global model, filtering out malicious or low-quality contributions. This includes \emph{Off-Chain Fairness Checks}, which aggregate contributions over multiple rounds, are performed off-chain to reduce blockchain computational overhead. Only the final fairness results are stored on-chain for transparency.
    \item \textbf{Aggregation:} The validated updates are aggregated using an efficient strategy to update the global model. We leverage the FedAvg algorithm~\cite{mcmahan2017communication}, which ensures that aggregation is computationally lightweight, reducing blockchain gas costs while maintaining model convergence.
    \item \textbf{Reward Distribution:} Rewards are allocated based on the alignment of updates, fairness evaluations, and participation consistency.
\end{itemize}

\item \textbf{Clients:}  
Participating clients perform local training on private datasets. They submit gradient updates to the smart contract while preserving data privacy, contributing to the global model without exposing sensitive information.

\end{itemize}

\section{Hybrid Incentive Mechanism Design}
\label{sec:incentive_mechanism}

\begin{figure}[t]
    \centering
    \includegraphics[width=\linewidth]{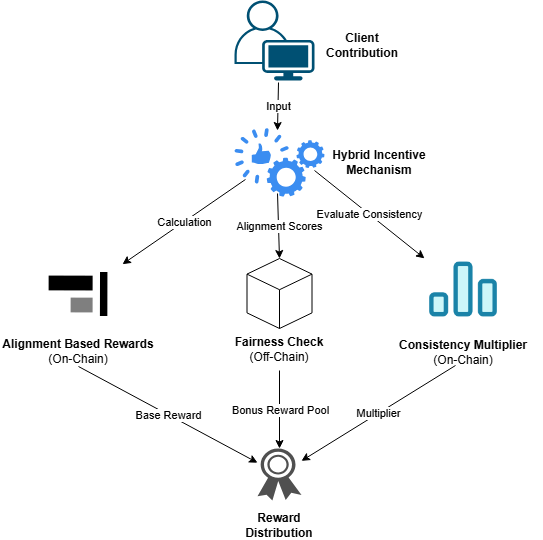}
    \Description{This figure depicts the hybrid incentive mechanism design for FL. It highlights the allocation of rewards through alignment-based rewards, periodic fairness checks, and consistency multipliers.}
    \caption{Allocation of Rewards.}
    \label{fig:hybrid_incentive_mechanism}
\end{figure}

\begin{figure*}[t]
    \centering
    \includegraphics[width=\linewidth]{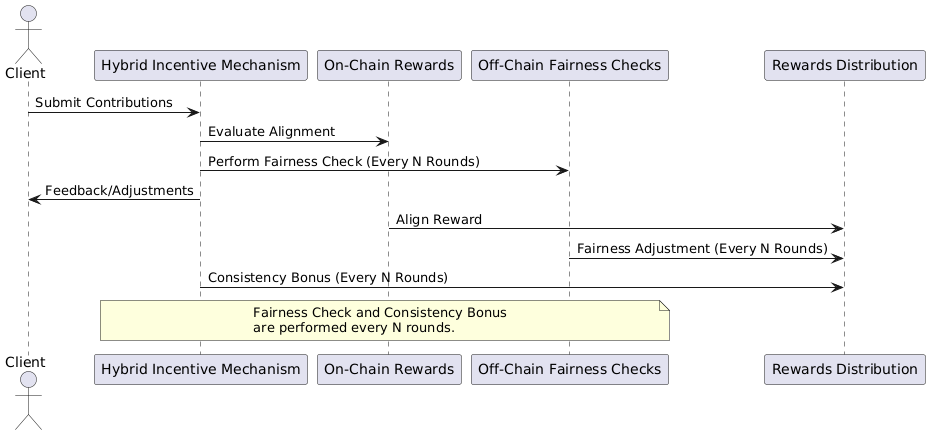}
    \Description{This figure shows a sequence diagram for the hybrid incentive mechanism. It details the flow of contributions, alignment-based rewards performed every round, and periodic fairness and consistency checks performed every N rounds.}
    \caption{Sequence Diagram for Hybrid Incentive Mechanism Design.}
    \label{fig:hybrid_incentive_sequence}
\end{figure*}

\begin{algorithm}
\caption{Hybrid Incentive Mechanism}
\label{alg:hybrid_incentive_mechanism}
\KwIn{Client updates \( \mathbf{g}_i \) for \( i \in \{1, 2, \dots, N\} \)}
\KwOut{Reward distribution for each client}

\SetAlgoLined
\textbf{Initialization:} Deploy smart contract; initialize global model \( \mathbf{g}_{\text{global}} \).

\ForEach{round $r$}{
    \textbf{Step 1: Update Submission:} \\
    \Indp Clients submit their gradient updates \( \mathbf{g}_i \) and metadata (e.g., \( n_i \)) to the smart contract.\\
    \Indm
    \textbf{Step 2: Alignment-Based Rewards:} \\
    \Indp Compute alignment scores \( S_i \) for each client \( i \) using:
    \[
    S_i = \left( \mathbf{g}_i \cdot \mathbf{g}_{\text{global}} \right) \cdot \frac{n_i}{N}
    \]
    Distribute rewards to clients with positive alignment scores.\\
    \Indm
    \textbf{Step 3: Model Aggregation:} \\
    \Indp Update the global model \( \mathbf{g}_{\text{global}} \) using FedAvg:
    \[
    \mathbf{g}_{\text{global}} = \sum_{i=1}^{N} \frac{n_i}{N} \mathbf{g}_i
    \]
    \Indm
}

\If{round $r$ is a multiple of fairness interval $N$}{
    \textbf{Step 4: Fairness Evaluation:} \\
    \Indp Compute cumulative scores \( C_i \) for each client \( i \):
    \[
    C_i = \sum_{r=1}^{R} S_{i,r}
    \]
    Store \( C_i \) values on IPFS and record the CID and integrity hash on the blockchain.\\
    \Indm
}

\textbf{Step 5: Reward Adjustment:} \\
\Indp Compute final rewards \( \text{Reward}_i \) for each client \( i \):
\[
\text{Reward}_i = S_i \cdot (1 + \alpha \cdot C)
\]
Distribute adjusted rewards to clients.\\
\Indm
\end{algorithm}

The hybrid incentive mechanism integrates three complementary components to address fairness, efficiency, and scalability challenges in decentralized FL systems as shown in \Cref{fig:hybrid_incentive_mechanism} and further illustrated in Algorithm \ref{alg:hybrid_incentive_mechanism}. 
These components ensure equitable rewards while promoting consistent and meaningful participation across multiple training rounds. 
\Cref{fig:hybrid_incentive_sequence} details the flow of contributions where the alignment-based rewards performed every round and periodic fairness and consistency checks performed every N rounds.


\subsection{Alignment-Based Rewards}
At the core of the incentive mechanism is the alignment-based reward system, which operates entirely on-chain. By leveraging smart contracts, the system ensures transparency and automation in evaluating client contributions. For each training round, the following steps are performed:
\begin{itemize}
    \item \textbf{Submission and Validation:} Clients submit updates (gradients) which are validated to filter out malicious or low-quality contributions.
    \item \textbf{Alignment Score Calculation:} Each client’s update is assessed based on its alignment with the aggregated global model direction.
\end{itemize}

The alignment score for client \(i\) is computed as:
\[
S_{i} = \left(\mathbf{g}_i \cdot \mathbf{g}_{\text{global}}\right) \cdot \frac{n_i}{N}
\]
where \(\mathbf{g}_i\) is the gradient submitted by client \(i\), \(\mathbf{g}_{\text{global}}\) is the aggregated gradient from all client updates, \(n_i\) is the number of data samples contributed by client \(i\), and \(N\) is the total number of data samples across all clients.

Using Shapley values~\cite{shapley1951notes}, we compute the contribution of client $i$ by evaluating their impact on the improvement of the global model across all possible subsets of clients. 
The Shapley value accounts for the improvement in model performance when client $i$ participates in a subset of clients, and the fairness of contributions, particularly for clients with smaller datasets or specialized data that might disproportionately benefit the global model.
Therefore, this adjustment ensures that rewards are distributed not only based on dataset size and alignment but also on the strategic importance of each client’s contribution to the overall training process. 

This above methodology ensures that clients are rewarded proportionally based on both the quality of their contributions and the scale of their data involvement. Misaligned or malicious updates are either penalized or ignored.
The alignment-based rewards are implemented via Solidity smart contracts, incorporating the following functionalities:
\begin{itemize}
    \item \textbf{Event Tracking:} Real-time updates on alignment scores through an \texttt{AlignmentScoresUpdated} event.
    \item \textbf{Reward Allocation:} Automated reward distribution to clients with positive alignment scores.
    \item \textbf{Gas Efficiency:} Computations are optimized to be lightweight to minimize on-chain resource usage.
\end{itemize}

\subsection{Periodic Fairness Checks via Cumulative Scoring}
To complement the short-term focus of alignment-based rewards, periodic fairness checks are conducted off-chain to ensure equitable distribution over longer periods. These checks operate as follows:
\begin{enumerate}[label=\arabic*.]
    \item \textbf{Cumulative Contribution Calculation:} Over \(R\) rounds, each client’s cumulative score \(C_i\) is computed as:
    \[
    C_i = \sum_{r=1}^{R} S_{i,r}
    \]
    where \(S_{i,r}\) is the alignment score for client \(i\) in round \(r\).
    
    \item \textbf{Data Storage and Transparency:} Cumulative scores are stored on the InterPlanetary File System (IPFS) for verifiability. The corresponding Content Identifier (CID) is recorded on-chain.
    
    \item \textbf{Integrity Validation:} A Keccak-256 hash of client contributions is generated for verification:
    \[
    H = \text{Keccak256}(C_i \ \forall i \in N)
    \]
    
    \item \textbf{On-Chain Updates:} The CID, cumulative scores, and integrity hash are updated on-chain via the smart contracts, as a tamper-proof record of fairness evaluations.
\end{enumerate}

The fairness checks bring several benefits:
\begin{itemize}
    \item \textbf{Transparency:} Verifiable cumulative data is accessible to all participants via IPFS.
    \item \textbf{Efficiency:} Off-chain computations reduce the blockchain’s computational load.
    \item \textbf{Fairness:} Aggregated rewards mitigate short-term biases, ensuring equitable treatment over time.
\end{itemize}

\subsection{Consistency Multipliers}
Consistency multipliers encourage sustained and meaningful participation by adjusting rewards based on client engagement across multiple rounds. The final reward for client \(i\) is calculated as:
\[
\text{Reward}_{i} = S_i \cdot (1 + \alpha \cdot C)
\]
where \(S_i\) is the alignment-based score of client \(i\), \(\alpha\) is the scaling factor for the consistency multiplier, and \(C\) is the proportion of rounds in which client \(i\) participated.

This formulation ensures that clients who consistently contribute high-quality updates are appropriately rewarded for long-term engagement.

\begin{table*}[t]
\centering
\caption{Gas Costs by Parameter Size}
\label{tab:gas_costs}
\begin{tabular}{|p{0.05\textwidth}|>{\raggedleft\arraybackslash}p{0.15\textwidth}|>{\raggedleft\arraybackslash}p{0.15\textwidth}|>{\raggedleft\arraybackslash}p{0.15\textwidth}|>{\raggedleft\arraybackslash}p{0.15\textwidth}|>{\raggedleft\arraybackslash}p{0.15\textwidth}|}
\hline
\textbf{Param Size} & \textbf{Registration \& Staking (per client)} & \textbf{Submit Model (per client)} & \textbf{Aggregate Models} & \textbf{Update Validation} & \textbf{Distribute Rewards} \\ \hline
10                  & 45,373                                          & 393,262                                & 499,660                       & 512,769                        & 219,961                         \\ \hline
100                 & 45,373                                          & 2,403,817                              & 3,891,311                     & 2,153,970                      & 219,961                         \\ \hline
1,000               & 45,373                                          & 22,866,722                             & 37,893,125                    & 18,609,485                     & 219,961                         \\ \hline
10,000              & 45,373                                          & 229,065,242                            & 386,438,410                   & 187,515,227                    & 219,961                         \\ \hline
100,000             & 45,373                                          & 2,447,670,138                          & 4,724,606,105                 & 2,311,631,243                  & 219,961                         \\ \hline
\end{tabular}
\end{table*}

\section{Evaluation}
\label{sec:evaluation}

Evaluating gas consumption in decentralized systems like blockchain-based FL quantifies the computational and storage costs of executing smart contract functions. Gas costs directly influence the economic feasibility and scalability of the system, particularly when scaling to large models and a high number of participants. We evaluated key smart contract operations, including registration, staking, model submission, aggregation, update validation, and reward distribution, under varying model parameter sizes. The results provide insights into the system’s efficiency and its applicability for resource-intensive tasks like training foundation models in federated settings.

\subsection{Gas Costs and Simulation Setup}

The relationship between gas units and actual dollar costs is determined by the gas price (in gwei), the gas price, and the current exchange rate of ETH to USD. 
Gas units also correlate with computational time, as they measure the complexity of smart contract operations; higher gas units typically indicate longer computation times due to more resource-intensive tasks. 

In this evaluation, we use the "Foundry" tool\footnote{\url{https://book.getfoundry.sh}} to deploy a private Ethereum blockchain. Since this is a simulation environment, we only report gas units as a proxy for computational costs. While these values provide an accurate estimate of resource requirements, they may differ from real-world deployments on the Ethereum mainnet due to variations in gas prices, network congestion, and token valuations. 

In practical settings, this framework is more likely to be deployed on a private Ethereum-based consortium blockchain, where the focus shifts from economic costs to computational efficiency as the primary metric for evaluating feasibility. This particular setup minimizes financial overhead while ensuring that the system’s performance meets the demands of real-world applications.

\subsection{Baseline Deployment Metrics}

The following deployment metrics establish a baseline for understanding the overhead associated with initializing the smart contract:
\begin{itemize}
    \item \textbf{Deployment Cost:} 2,371,244 gas
    \item \textbf{Deployment Size:} 10,667 bytes
\end{itemize}

\subsection{Gas Consumption Analysis}

The gas consumption results in \Cref{tab:gas_costs} highlight the scalability challenges of aggregating large models entirely on-chain in blockchain-based FL settings. For smaller models, such as those with 10 or 100 parameters, gas costs for operations like \emph{Submit Models}, \emph{Aggregate Models}, and \emph{Update Validation} are manageable. However, as the model size increases, gas costs grow exponentially, making such operations impractical for models with 100,000 parameters or more. This exponential growth underscores the limitations of on-chain processing for large-scale FL.

In contrast, operations such as \emph{Registration \& Staking} and \emph{Distribute Rewards} incur constant gas costs regardless of the model size. This is because these operations are independent of the parameter size and are efficiently implemented using smart contracts.

The results demonstrate that while on-chain operations are feasible for small to moderately sized models, alternative approaches like batch processing and off-chain computations are essential for scaling to resource-intensive models like LLMs. These optimizations ensure that blockchain-based FL can remain cost-effective and efficient while supporting the training of high-complexity models in decentralized environments.




\section{Discussion}
\label{sec:discussion}

Our findings underscore the importance of aligning model size with the capabilities of the underlying blockchain infrastructure to achieve practical and scalable FL. Given the observed scalability limitations as noted in \Cref{tab:gas_costs}, smaller LLMs may be better suited for FL settings when using on-chain aggregation. 

\subsection{Limitations}

Although the current evaluation offers a comprehensive analysis of gas costs associated with deploying, training, and updating the proposed blockchain-based federated learning framework, it does not fully address the performance evaluation of the incentive and reward mechanisms. Specifically, the on-chain alignment-based rewards, off-chain fairness checks, and consistency multipliers—key components of the framework—remain unevaluated in terms of their impact on fairness, transparency, and participant engagement. These mechanisms play a critical role in incentivizing meaningful contributions and maintaining long-term system participation, particularly in heterogeneous and competitive environments. A thorough analysis of their operational efficiency, scalability, and effectiveness is essential to assess the framework's overall feasibility in real-world deployments. 

\subsection{Challenges and Mitigation Strategies}

Building on the identified limitations, this section outlines key challenges faced by the proposed blockchain-based FL framework and the mitigation strategies designed to address them. These challenges, including scalability, heterogeneous data distributions, and alignment gaming, require tailored solutions especially to ensure the framework’s efficiency, fairness, and robustness in real-world deployments.

\subsubsection{Scalability}
\begin{itemize}
    \item \textbf{Batch Processing:} Model submissions and computations can be split into smaller chunks, such as 10,000 parameters per batch. This mitigates the exponential increase in gas consumption for large models while preserving the integrity of the aggregation process.
    
    \item \textbf{Off-Chain Aggregation and Validation:} Performing resource-intensive operations like aggregation and update validation off-chain significantly reduces gas costs. Only the final results, such as cryptographic hashes, are submitted on-chain for verification. This approach ensures computational efficiency while maintaining the trust and transparency inherent in blockchain-based systems.
    
    \item \textbf{Efficient Aggregation Techniques for Large Networks:} Employing advanced aggregation methods, such as sparse updates or quantized model representations, can significantly reduce computational overhead and the volume of data transmitted. These techniques ensure that the framework can scale effectively while maintaining model performance.
\end{itemize}

\subsubsection{Heterogeneous Data}
\begin{itemize}
    \item \textbf{Weighted Alignment Scores:} To address biases caused by varying dataset sizes among clients, alignment scores are weighted based on the sample size contributed by each client. This ensures that smaller datasets do not disproportionately affect the global model's training process.
    
    \item \textbf{Periodic Fairness Checks:} Specialized contributors with unique or highly valuable data may risk being under-rewarded in conventional systems. Periodic fairness checks ensure equitable reward distribution by assessing long-term contributions and adjusting for any imbalances.
\end{itemize}

\subsubsection{Alignment Gaming}
\begin{itemize}
    \item \textbf{Penalty for Negative Alignment Scores:} To discourage malicious behavior, updates with consistently negative alignment scores are penalized~\cite{zhao2023}. This ensures that clients who provide harmful or low-quality updates are deterred from gaming the system.
    
    \item \textbf{Accuracy-Adjusted Alignment Metrics:} The alignment metrics are refined to incorporate improvements in model accuracy. This adjustment ensures that client contributions are evaluated not only on alignment with the global gradient but also on their impact on the model's overall performance.
\end{itemize}



\subsection{Future Work}

Future efforts will focus on addressing several critical aspects to further enhance the proposed blockchain-based FL framework.

First, optimizing computational resources for LLMs and other models with large parameter settings is a key priority. This can be achieved by offloading large parameter aggregation to decentralized storage solutions, such as IPFS, in trustful settings, thereby reducing the computational and storage burden on the blockchain. However, when such models must be aggregated on a public permissionless blockchain, additional strategies are required to address scalability and security concerns.
In such scenarios, aggregation can be performed in a batched manner, where only hashed summaries or intermediate results of smaller parameter chunks are recorded on-chain. This reduces the size of on-chain transactions while preserving transparency and verifiability. 

Second, improving reward scaling mechanisms is essential to ensure fairness and efficiency across a wide range of FL scenarios. 
real-world FL scenarios with heterogeneous data distributions and client behaviors.
This includes comprehensive testing in environments with diverse client behaviors and varying data contributions to address discrepancies in reward distribution.
We plan to simulate real-world FL scenarios with diverse client behaviors and measuring the gas costs and computational overhead associated with each incentive mechanism. Additionally, empirical studies could explore the trade-offs between the economic costs of running these mechanisms and their impact on participation rates, fairness, and model performance, providing a more comprehensive assessment of the framework. This expanded evaluation would further validate the practicality of the proposed hybrid incentive mechanism and identify optimization opportunities for large-scale deployments.

Finally, establishing a standardized benchmark for evaluating FL architectures integrated with the proposed mechanism will be valuable for comparing different architectures and guiding future optimizations.

\section{Conclusion}
\label{sec:conclusion}

This paper introduces a blockchain-based framework that tackles key challenges in traditional FL systems, including trust, fairness, and scalability. By leveraging blockchain technology and smart contracts, the framework automates critical operations such as client registration, update validation, reward distribution, and global state maintenance, eliminating the need for centralized aggregation mechanisms. The hybrid incentive mechanism—combining on-chain alignment-based rewards, off-chain fairness checks, and consistency multipliers—ensures equitable participation, sustained engagement, and efficient resource utilization, making the framework robust and adaptable to heterogeneous environments.

Empirical evaluations validate the framework’s feasibility, demonstrating that it is well-suited for decentralized and collaborative training scenarios, particularly for models with moderate parameter sizes. The gas efficiency results underscore the practicality of the framework in addressing the computational and economic constraints associated with blockchain-based systems. 

As machine learning continues to scale, with foundation models and other resource-intensive architectures at the forefront, the framework proposed in this paper offers a promising path forward by enabling equitable participation and contributions in decentralized training. Finally, by addressing core challenges in trust, scalability, and fairness, this work represents a critical step toward realizing efficient, decentralized AI solutions that are both practical and impactful for real-world applications.



\section*{Code Availability}

The source code for the blockchain-based FL aggregator, including the Solidity smart contracts and Python scripts for off-chain computations, is available on our GitHub repository at \url{https://github.com/brains-group/OpenFedLLM/tree/Bijun-SmartContract/smart_contract}.



\balance

\bibliographystyle{ACM-Reference-Format}
\bibliography{reference}



\end{document}